\documentclass{article}

\usepackage[utf8]{inputenc} 
\usepackage[T1]{fontenc}    
\usepackage{hyperref}       
\usepackage{url}            
\usepackage{booktabs}       
\usepackage{amsfonts}       
\usepackage{nicefrac}       
\usepackage{microtype}      
\usepackage{xcolor}         
\pdfoutput=1
\usepackage{yfonts}
\usepackage{amssymb} 
\usepackage{mathtools}
\usepackage{epsfig} 
\usepackage{float} 
\usepackage[caption = false]{subfig}
\usepackage{wasysym} 
\usepackage{mathrsfs} 
\usepackage{amsfonts} 
\usepackage{amsbsy} 
\usepackage{amscd} 
\usepackage{pstricks} 
\usepackage{multirow} 
\usepackage{tikz}
\usepackage{color}
\usepackage{slashed}
\usepackage{array}
\usepackage{tablefootnote}

\usetikzlibrary{arrows,positioning,shapes.geometric} 
\usepackage[font=small,labelfont=bf]{caption}
\usepackage{slashed}
\usepackage{multirow}
\usepackage{tabularx}
\usepackage{soul}  
\usepackage{listings} 
\usepackage[margin=1in]{geometry}
\usepackage{color}

\usepackage{amsthm}

\newtheorem{definition}{Definition}

\newtheorem{observation}{Observation}

\newcommand{\im}{\text{Im}} 

\newcommand{\mbf}[1]{\mathbf{#1}}

\title{
	 Optimal Symmetries in Binary Classification 
	} 
\author{%
	Vishal~S.~Ngairangbam$^{1}$\footnote{email: vishal.ngairangbam@kit.edu, corresponding Author},\hspace{10pt}and Michael~Spannowsky$^{1,2}$\footnote{email: michael.spannowsky@kit.edu}\; \\
	$^{1}$Department of Physics, Durham University\\
	Durham DH1 3LE, United Kingdom\\
	$^{2}$Institute for Theoretical Physics, Karlsruhe Institute of Technology, \\ Wolfgang-Gaede-Str. 1, 76131 Karlsruhe, Germany	
}

\allowdisplaybreaks
\begin{document} 
	\maketitle
\begin{abstract}
	We develop a theoretical foundation for designing group-equivariant neural networks that align the choice of symmetries with the underlying probability distributions of the data.  Utilising the general structure of fibre decompositions on the domain under group equivariant maps and its relation to that of the likelihood ratio, we present a theoretical framework for identifying group actions that  maintain optimal classification performance via the Neyman-Pearson lemma. 
	This provides a unified methodology for improving classification accuracy especially in fundamental applications where one has knowledge of the inherent symmetries of the distributions and how they are broken by measurement.	As an application to jet classification at the Large Hadron Collider, we find that there can be performance gains when one utilises smaller permutation symmetries within the constituents. This work offers insights and practical guidelines for constructing more effective group equivariant architectures in diverse machine-learning contexts.
\end{abstract} 	
\section{Introduction}

Equivariant neural networks, which leverage symmetries inherent in data, have shown significant promise in enhancing classification accuracy, data efficiency, and convergence speed. However, selecting an appropriate group and defining its actions across the various layers of the network remains a complex task, particularly for applications requiring adherence to specific symmetries. This research aims to establish a foundational framework for designing equivariant neural network architectures by utilising stabiliser groups.

In the context of equivariant function approximation, a critical insight is that the preimage of a target element in the output space can be characterised by the subset of the orbit of an input element, restricted to those connected by elements of the stabiliser group of the target. This understanding simplifies the choice of group $\mathcal{G}$ and its actions, aligning them with the equivalence class of the target function’s fibres in the input space.

This paper provides a theoretical basis for designing equivariant architectures in scientific applications where known symmetries play a crucial role. By grounding the architecture design in first principles, we aim to explain empirical successes observed in data-driven applications and enhance the effectiveness of these architectures in practical scenarios.

\paragraph{Our contribution:} In this work, we utilise the minimal fibres that any group invariant or equivariant function should have and connect it to the optimality of a particular symmetry. Although the result is a simple extension of the well-known Neyman Pearson lemma~\cite{doi:10.1098/rsta.1933.0009} where an optimal binary classifier should have the same fibre decomposition in the domain of the two probabilities as the likelihood ratio, it provides a transparent basis for maximally utilising domain-knowledge in simulation-based inference, which is quickly gaining popularity in various disciplines. 
Defining a compatible symmetry as one that induces compatible fibres on the domain with a given likelihood ratio, we find that:
\begin{itemize}
	\item if the support of the two probabilities contain an entire orbit of a compatible symmetry, then equivariant functions under any non-trivial action on the co-domain of group that is contained in such a symmetry has incompatible fibres with the likelihood ratio (see observation~\ref{obs:no_go_co_domain}) 
	\item there always exists a maximal symmetric action of a group $\mathcal{G}_0$ after which one can include non-trivial equivariant $\mathcal{G}$-actions groups that contain the $\mathcal{G}_0$-action as a proper subgroup restriction as long as the corresponding $\mathcal{G}$-action of group element does not belong to $\mathcal{G}_0$ takes any element out of the support (see observation~\ref{obs:equi_opt})
	\item As a practical application, we analyse the optimality of permutation symmetries in jet classification at the Large Hadron Collider and find that permutation over the entire jet constituents can be suboptimal especially, near the resonant mass beak of the boosted resonant particles. This is substantiated by numerical experiments where we find differences improvements in performance when one utilises the bottom quark information in top classification. Moreover, this difference increases when one reduces the mass window of the selection criterion. Similarly, we also verify the improvement in quark/gluon classification by reducing the permutation symmetries via the particle's charge information.  
\end{itemize}

\section{Related works}

\paragraph{Group Equivariant Networks:} Group equivariant architecture design is ubiquitous in modern deep learning algorithms with Convolutional Neural Networks~\cite{10.1162/neco.1989.1.4.541} being equivariant to translations.  Modern generalisations include equivariance to discrete groups~\cite{pmlr-v48-cohenc16}, and gauge symmetries~\cite{pmlr-v97-cohen19d}.  It was shown in~\cite{pmlr-v80-kondor18a} that equivariance with respect to any compact group requires the operation of a layer to be equivalent to a generalised convolution on the group elements. 
General frameworks for Lie group equivariance have developed  in~\cite{pmlr-v119-finzi20a,Bekkers2020B-Spline,pmlr-v139-hutchinson21a}.

For set-based approaches, the DeepSets framework~\cite{NIPS2017_f22e4747} formulated the universal structure of permutation equivariant and invariant functions on elements belonging to a countable universe. Equivariant feature extraction on point clouds for the classifical groups has also been explored in~\cite{Chen_2021_CVPR,pmlr-v139-satorras21a,villar2021scalars}. 
Geometric Algebra Transformer~\cite{NEURIPS2023_6f6dd92b,pmlr-v238-haan24a}  have been designed for 3D equivariant extraction of various non-trivial transformation groups, including the conformal and projective groups~\cite{pmlr-v238-haan24a} utilising Clifford group equivariant  networks~\cite{ruhe2023clifford}.  
\paragraph{Theoretical analysis of equivariant models:} On top of the empirical results, several theoretical results exist on the advantages of equivariant models. Under the assumption that the classification task is invariant to symmetry transformation, bounds on the generalisation error were derived in~\cite{pmlr-v54-sokolic17a}.  The sample complexity for invariant kernel methods is studied in~\cite{pmlr-v134-mei21a}.   In the framework of PAC learning, it was shown~\cite{elesedy2022group} that learning with invariant or equivariant methods reduces to learning over orbit representatives. At the same time, the incorporation of appropriate group equivariances~\cite{pmlr-v139-elesedy21a,NEURIPS2021_8fe04df4} results in a strict increase in generalisation. The general intuition from these works is that larger groups result in better sample efficiency and improved generalisation when the target function follows the assumed symmetry.  Our work contributes in this direction by defining the notion of a correct symmetry for a binary classification task with a priori known symmetries of the probabilities.

\paragraph{Equivariance in fundamental physics:}  The importance of equivariance in fundamental physics cannot be overstated, starting from the Galilean transformations in non-relativistic regime to the exotic structure of gauge theories, which form the mathematical backbone of our current understanding of the universe. Lorentz group equivariance has been explored in~\cite{pmlr-v119-bogatskiy20a,Gong:2022lye,Bogatskiy:2023nnw,Spinner:2024hjm}. Other group equivariance studies include architectures for simulating QCD on the lattice~\cite{PhysRevLett.128.032003}, and the classification of radio galaxies~\cite{10.1093/mnras/stab530}.

\section{Preliminaries} 
\paragraph{Fibres of a function:} Our work relies on studying the minimal fibre structure induced by symmetries on the likelihood ratio and its relation to different group equivariant functions. A fibre is formally defined as:  
\begin{definition}
	Given a function $f:\mathcal{X}\to\mathcal{Y}$, the fibre of an element $\mathbf{y}$ in the image of $f$ denoted as $f^{-1}(\mathbf{y})$, is the set of all elements of $\mathcal{X}$ which gets mapped to $\mathbf{y}$ via $f$: $$f^{-1}(\mathbf{y})=\{\mathbf{x}\in\mathcal{X}\;:\; \mathbf{f}(\mathbf{x})=\mathbf{y}\}\quad.$$ 
\end{definition}The fibres of all element in $\im(f)$ divide $\mathcal{X}$ into equivalence classes.   This means that the set of fibres of any function partitions the domain $\mathcal{D}$ into mutually exclusive subsets. This holds true for any equivalence class in any given set, i.e. any element $\mathbf{x}\in\mathcal{X}$ can only belong to a single equivalence class. We utilise this basic fact implicitly in different parts of the work. 
\paragraph{Group actions:} The main idea underlying symmetries is the action of groups on an underlying set. Group actions and some related terminologies are defined as follows. 
\begin{definition}
	Given a group $\mathcal{G}$ and a set $\mathcal{X}$, the left-action of $\mathcal{G}$ on $\mathcal{X}$ is a  map $\mbf{A}:\mathcal{G} \times \mathcal{X} \to \mathcal{X}$ with the following properties: 
	\begin{enumerate}
		\item  $\mbf{A}(e,\mbf{X})=\mbf{X}$ for the identity element $e\in\mathcal{G}$ and any $\mbf{X}\in\mathcal{X}$, and   
		\item  $\mbf{A}(g_1,\mbf{A}(g_2,\mbf{X}))=\mbf{A}(g_1g_2,\mbf{X})$  for all $g_1,g_2\in\mathcal{G}$ and $\mathbf{X}\in\mathcal{X}$. 
	\end{enumerate}   
\end{definition} 
A $\mathcal{G}$-action will be denoted as $(\mathcal{G},\mathbf{A}(g,\mathbf{X}))$. Any set always admits the trivial action where $\mathbf{A}(g,\mathbf{X})=\mathbf{X}$ for any group element $g$ and all elements $\mathbf{X}$ of the set $\mathcal{X}$.  
Consequently, even for the same group $\mathcal{G}$, one can have two actions $\mathbf{A}_1(g,\mathbf{X})$ and $\mathbf{A}_2(g,\mathbf{X})$ acting on the same set $\mathcal{X}$, which are not necessarily the same functions. This motivates defining the equality of two $\mathcal{G}$-actions as follows. 
\begin{definition}
	\label{def:act_equal} 
	Two $\mathcal{G}$-actions $(\mathcal{G}_1,\mathbf{A}_1(g,\mathbf{X}))$ and $(\mathcal{G}_2,\mathbf{A}_2(g,\mathbf{X}))$ acting on the same set $\mathcal{X}\ni\mathbf{X}$ are equal if and only if $\mathcal{G}_1$ is isomorphic to $\mathcal{G}_2$, via an invertible map $i:\mathcal{G}_1\to\mathcal{G}_2$ and $\mathbf{A}_1(g,\mathbf{X})=\mathbf{A}_2(i(g),\mathbf{X})$ for all $g\in\mathcal{G}_1$ and $\mathbf{X}\in\mathcal{X}$. 
\end{definition} 
We will primarily concern ourselves with faithful group actions. They are defined as:
\begin{definition}
	A $\mathcal{G}$-action $\mathbf{A}(g,\mathbf{X})$ on a set $\mathbf{X}$ is faithful if $A(g,\mathbf{X})=\mathbf{X}$ for all $\mathbf{X}\in\mathcal{X}$ then $g=e$. 
\end{definition}
Primarily, we will be utilising the property of a faithful action that any group element $g$ moves at least one element $\mathbf{X}_g$ of the set $\mathcal{X}$.

\begin{definition} 
	The orbit of an element $\mbf{X}\in\mathcal{X}$ under the group action $\mbf{A}(g,\mbf{X})$ of the group $\mathcal{G}$, is the set of all elements $\mbf{X}'\in\mathcal{X}$ for which there is a group element $g\in\mathcal{G}$ such that $\mbf{X}'=\mbf{A}(g,\mbf{X})$. 
\end{definition}
We will write the orbit of a $\mathcal{G}$-action characterised by a vector of group invariants $\mathbf{I}$ as 
\begin{equation}
	\Omega^\mathcal{X}(\mathbf{I})=\{\mathbf{X}'=\mbf{A}(g,\mathbf{X}) :\{\mathbf{X},\mathbf{X}'\}\subset\mathcal{X} \text{ and } g\in\mathcal{G}\}\quad.  
\end{equation} The set of all distinct orbits divides $\mathcal{X}$ into equivalence classes. 
\begin{definition}
	The stabiliser group of an element $\mathbf{X}\in\mathcal{X}$ with respect to the $\mathcal{G}$-action $\mathbf{A}$, is the set of all group elements $g$ which leave $\mathbf{X}$ invariant.  Mathematically, we will denote it as 
	\begin{equation*}
		\mathcal{L}^\mathbf{A}(\mathbf{X})=\{g\in\mathcal{G} : \mathbf{X}=\mathbf{A}(g,\mathbf{X})\}\subseteq \mathcal{G} 
	\end{equation*}
\end{definition}
A group representation $\rho:\mathcal{G}\to GL(m,\mathbb{R})$, naturally induces a linear action $\mathbf{A}_\rho:\mathcal{G}\times\mathbb{R}^m\to\mathbb{R}^m$ on $\mathbb{R}^m$ of the form $\mathbf{A}_\rho(g,\mathbf{x})=\rho(g)\,\mathbf{x}$. In the rest of the paper, we will be primarily concerned with such actions of a given transformation group on $\mathbb{R}^m$.  We will work under the relaxation that these actions need not be closed in the domain of the target function  $\mathcal{D}\subset\mathbb{R}^m$, i.e. there may be group elements $g\in\mathcal{G}$ which takes a point $\mathbf{x}\in\mathcal{D}$ to $\rho(g)\,\mathbf{x}=\mathbf{x}'\notin\mathcal{D}$. For a pedagogical overview of common transformation groups see~\cite{WEYL1966}. 

\paragraph{Group equivariant functions:} Symmetries are built into neural networks using group equivariant maps, defined as follows. 
\begin{definition}
	A function $\mbf{f}:\mathcal{X}\to\mathcal{H}$ between two spaces $\mathcal{X}$ and $\mathcal{H}$ which both admit group actions, say $\mbf{A}_\mathcal{X}$ and $\mbf{A}_\mathcal{H}$ respectively, for a group $\mathcal{G}$ is said to be $\mathcal{G}$-equivariant if $\mbf{f}$ commutes with the group actions, i.e., 
	\begin{equation}
		\label{eq:equiv} 
		\mbf{f}(\mbf{A}_\mathcal{X}(g,\mbf{X}))=\mbf{A}_\mathcal{H}(g,\mbf{f}(\mbf{X})) \quad,
	\end{equation}for every $g\in\mathcal{G}$ and $\mbf{X}\in\mathcal{X}$.  The map $\mbf{f}$ is $\mathcal{G}$-invariant if the action $\mbf{A}_\mathcal{H}$ is trivial, i.e., $\mbf{A}_\mathcal{H}(g,\mbf{H})=\mbf{H}$ for all $g\in\mathcal{G}$ and any $\mbf{H}\in \mathcal{H}$.  
\end{definition}

\paragraph{Likelihood ratio and hypothesis testing:}  
We first outline a non-technical and intuitive picture of the ubiquitous Neyman-Pearson lemma~\cite{doi:10.1098/rsta.1933.0009} which we will then connect to optimal binary classification. Consider an observed data $\mathbf{X}$ and we hypothesise that it originates from either $P_0(\mathbf{X},\Theta_0)$ or $P_1(\mathbf{X},\Theta_1)$, where $\Theta_\alpha$ are parameters of the probability distribution. Those hypotheses where the probabilities are completely specified are said to be \emph{simple}, i.e., each of the parameters $\Theta_\alpha$ are fixed to specific values for $\alpha\in\{0,1\}$. The Neyman-Pearson lemma applies to scenarios where the two hypotheses are simple and exhaustive, i.e., the data can originate from only these two completely specified probabilities. The null hypothesis $H_0$, where $\mathbf{X}$ follows $P_0$, is usually chosen to represent the currently understood phenomena, while the alternate hypothesis $H_1$, where $\mathbf{X}$ follows $P_1$, are taken to describe possible phenomena in the new regime probed by the experiment.

The power of a statistical test is the probability of correctly rejecting the null hypothesis when the alternate hypothesis is true. At the same time, the significance is the probability of rejecting the null hypothesis when it is true. The Neyman-Pearson lemma states that the likelihood ratio $\lambda(\mathbf{X})=P_1(\mathbf{X})/P_0(\mathbf{X})$ is uniformly the most powerful test statistic via which we can accept that $\mathbf{X}$ originated from $P_1$ for a given significance. Notably, a binary classification problem with an appropriate loss function (including binary cross-entropy) reduces to approximating a monotonic function of the likelihood ratio~\cite{Cranmer:2015bka}.

\paragraph{Optimal binary classification:}  
As we aim to connect binary classification with the Neyman-Pearson optimality of hypothesis tests, we will utilise an analogous and straightforward definition of optimality in binary classification following the standard formulation of the receiver operator characteristics curve. 
Let us take the binary classification of a sample $\mathbf{X}$ sampled from either $P_0(\mathbf{X})$ or $P_1(\mathbf{X})$ with the same support $\mathcal{D}\subset\mathbb{R}^m$, where we are interested in maximally selecting those samples originating from $P_1$. Often, the support of the alternate hypothesis' probability distribution function (pdf) $P_1$ is contained within that of the null hypothesis $P_0$.  On the other hand, if there is some part of $P_1$'s support outside $P_0$, any event obtained in this region trivially can only follow the alternate hypothesis. We are, therefore, primarily interested in those non-trivial cases where the data falls in the intersection of the support of $P_0$ and $P_1$. For a given classifier $\hat{f}:\mathcal{D}\to \mathbb{R}$ which learns the underlying class assignment map by segregating samples from $P_1$ to negative values, dependent on a threshold $t\in \mathbb{R}$ the cumulative distribution of $\hat{y}=\hat{f}(\mathbf{X})$ under $P_\alpha$ are 
$$\epsilon_\alpha=C_\alpha(t)=\int_{\mathcal{D}} dV\, P_\alpha(\mathbf{X})\;\Theta(\hat{y}<t)\quad,$$ where $dV$ is the volume element in $\mathcal{D}$.  Therefore, $\epsilon_1$ is the true positive rate and $\epsilon_0$ is the false positive rate and the former can be cast as a function of the later using the threshold $t$, $\epsilon_1=C_1\circ C^{-1}_0(\epsilon_0)=\text{ROC}(\epsilon_0)$. We now have the following definition of an optimal binary classifier. 
\begin{definition} 
	\label{def:opt_perf} 
	A binary classifier for data sampled from two distributions $P_0(\mathbf{x})$ and $P_1(\mathbf{x})$ is optimal in accepting samples from $P_1$ at a given tolerance $\epsilon_0$ of accepting false samples from $P_0$, if it has the highest possible acceptance of samples $\epsilon_1$ from $P_1$.  
\end{definition}  
The connection to Neyman-Pearson's optimality of the likelihood ratio can be seen as follows. For each $\mathbf{X}$, if the null hypothesis $H_0$ is that the underlying distribution is $P_0$ and the alternate hypothesis $H_1$ is that it is  $P_1$, it is straightforward to see that $\epsilon_0$ is the significance of the test and $\epsilon_1$ is its power. Therefore, we have the following observation.

\begin{observation}
	\label{obs:opt_class} 
	A binary classifier $\hat{f}:\mathcal{D}\to\mathbb{R}$ of data $\mathbf{X}\in\mathcal{D}$ sampled from either $P_0(\mathbf{X})$ or $P_1(\mathbf{X})$ is an optimal classifier for all values of $\epsilon_0$, if and only if it is a monotonic function of the likelihood ratio $P_1(\mathbf{X})/P_0(\mathbf{X})$. 
\end{observation} 
The relaxation to a monotonic function of the likelihood ratio follows from the integral definition of $\epsilon_\alpha$ depending implicitly on the threshold, i.e., we are interested in choosing regions in $\mathcal{D}$ divided by hypersurfaces which have constant likelihood ratios. For a $d$-dimensional domain, these are at least $d-1$ dimensional, and therefore, they are computationally non-trivial to evaluate in high dimensions even when one can write down closed-form expressions for the probabilities. Additionally, a necessary condition for any function to be the optimal classifier is to have an identical fibre decomposition in the domain $\mathcal{D}$ with the likelihood ratio. 

\section{Minimal fibres of group equivariant functions}
\label{sec:fibre}  
Since a necessary optimality condition in binary classification is to have the same fibre decomposition in the domain $\mathcal{D}$, we now describe the minimal fibres of group of equivariant functions. 

\subsection{Invariant functions} 
Let $\mathbf{h}:\mathcal{X}\to\mathcal{H}$ be a $\mathcal{G}$-invariant function, i.e. for any $\mathbf{X}\in\mathcal{X}$ and all $g\in\mathcal{G}$, \begin{equation}
	\label{eq:g_inv} 
	\mathbf{h}(\mathbf{A}_\mathcal{X}(g,\mathbf{X})) =\mathbf{h}(\mathbf{X})\quad.  
\end{equation} 

This means that if  $\mathbf{X}\in\Omega^{\mathcal{X}}(\mathbf{I})$ then for all $\mathbf{X}'=\mathbf{A}_\mathcal{X}(g,\mathbf{X})\in\Omega^\mathcal{X}(\mathbf{I})$,  $\mathbf{h}(\mathbf{X}')=\mathbf{h}(\mathbf{X})$.  Therefore, we have the following observation:
\begin{observation} 
	\label{obs:inv_fibre} 
	The fibre of an element $\mathbf{H}=\mathbf{h}(\mathbf{X})\in\mathcal{H}$ in the image of a $\mathcal{G}$-invariant function $\mathbf{h}:\mathcal{X}\to\mathcal{H}$ is at least as large as the orbit $\Omega^{\mathcal{X}}(\mathbf{I})$ of $\mathbf{X}\in\mathcal{X}$ of the action $\mathbf{A}_\mathcal{X}$. If the map is equal for elements belonging to distinct orbits, the fibre becomes enlarged to the union of these orbits.   
\end{observation} 
Let $\mathbf{h}(\mathbf{X})$ become equal for $\mathbf{X}$ in all orbits parametrized by $\mathbf{I}$ in the set $\mathcal{F}$ of orbit invariants $\mathbf{I}$. The fibre $\mathbf{h}(\mathbf{X})$ for any $\mathbf{X}$ in these orbits is        
\begin{equation} 
	\mathbf{h}^{-1}(\mathbf{X})=\bigcup_{\mathbf{I}\in\mathcal{F}} \Omega^\mathcal{X}(\mathbf{I})\quad.
\end{equation}

\subsection{Equivariant functions}  
Let  $\mathbf{f}:\mathcal{X}\to\mathcal{H}$ be a $\mathcal{G}$-equivariant function with respect to the actions $\mathbf{A}_\mathcal{X}$ and $\mathbf{A}_\mathcal{H}$ on $\mathcal{X}$ and $\mathcal{H}$, respectively. If $\mathbf{X}'=\mathbf{A}_\mathcal{X}(g,\mathbf{X})$ is in the orbit of $\mathbf{X}$, from Eq~\ref{eq:equiv}, we have $\mathbf{f}(\mathbf{X}')=\mathbf{A}_\mathcal{H}(g,\mathbf{f}(\mathbf{X}))$. It is straightforward to see that 
\begin{equation} 
	g\in\mathcal{L}^{\mathbf{A}_\mathcal{H}}(\mathbf{f}(\mathbf{X}))\implies \mathbf{f}(\mathbf{X}')=\mathbf{f}(\mathbf{X})\quad,
\end{equation}  i.e., if $\mathbf{X}$ and $\mathbf{X}'$ are connected by a group element $g$ which belongs to the stabiliser group of the action $\mathbf{A}_\mathcal{H}$ at $\mathbf{f}(\mathbf{X})$, then $\mathbf{X}'$ is contained in the fibre $\mathbf{f}^{-1}(\mathbf{H})$ of $\mathbf{H}=\mathbf{f}(\mathbf{X})$. 
Therefore, similar to the case for invariant maps, we have the following observation: 
\begin{observation}
	\label{obs:equi_fibre} 
	\textit{The fibre of an element $\mathbf{H}=\mathbf{f}(\mathbf{X})\in\mathcal{H}$ in the image of a $\mathcal{G}$-equivariant function $\mathbf{f}:\mathcal{X}\to\mathcal{H}$, is at least as large as the subset of the orbit of $\mathbf{X}$ connected by an element $g$ in the stabiliser group of $\mathbf{A}_\mathcal{H}$ at $\mathbf{H}$. If the map is equal for elements belonging to distinct orbits in $\mathcal{X}$, the fibre becomes enlarged to the union of all such sets. 
	} 
\end{observation}
Define the subset of the orbit $\Omega^\mathcal{X}(\mathbf{I})$ dependent on $\mathbf{H}=\mathbf{f}(\mathbf{X})$ as  \begin{equation} 
	\label{eq:equiv_orbit}
	\mathcal{O}(\mathbf{I},\mathbf{H})=\{\mathbf{X}'=\mathbf{A}_\mathcal{X}(g,\mathbf{X}) : g\in\mathcal{L}^{\mathbf{A}_\mathcal{H}}(\mathbf{H})\}\quad.
\end{equation} 
If $\mathbf{f}(\mathbf{X})$ become equal for $\mathbf{X}$ in all such subsets of distinct orbits parametrized by $\mathbf{I}$ in the subset $\mathcal{F}\subseteq\mathcal{I}$. 
The fibres can be written as 
\begin{equation}
	\label{eq:equi_fibre} 
	\mathbf{f}^{-1}(\mathbf{H})=\bigcup_{\mathbf{I}\in\mathcal{F}} \, \mathcal{O}(\mathbf{I},\mathbf{H}) \quad. 
\end{equation} 
As expected, the invariant result is a special case when the action $\mathbf{A}_\mathcal{H}$ is trivial, i.e. the stabiliser group of any element $\mathbf{H}\in\mathcal{H}$ is the group $\mathcal{G}$ and $\mathcal{O}(\mathbf{I},\mathbf{H})=\Omega^\mathcal{X}(\mathbf{I})$ for any $\mathbf{H}$.

As we will be primarily concerned with probabilities with support on a subset $\mathcal{D}\subset\mathcal{X}$ where  $\mathcal{X}=\mathbb{R}^m$, the fibres become the intersection of the respective sets with the domain $\mathcal{D}$. At this point, we point out that for general equivariant function approximation, a correct symmetry (whether invariant or equivariant) is one which does not mix the fibres of a target function through the minimal fibres induced via group equivariance.

\section{Optimal symmetries in binary classification} 
In fundamental applications, more often than not, the probabilities are invariant under some transformation group action on the domain $\mathcal{D}$. Even when closed-form expressions are not known, various first-principle arguments require the probabilities to be invariant under a transformation group. In this section, we answer the question of which symmetries retain the optimality of a classifier when the symmetry of the two probabilities is known a priori. We ignore the effects of noise in our mathematical description while we observe that the findings persist in the presence of noise in our numerical experiments.

\subsection{Compatible, closed, and maximal group symmetries} 
Definition~\ref{def:opt_perf} considers optimality in terms of performance alone. To connect with the structure of fibres that some assumed group equivariance endows on the domain, we first define a correct symmetry for binary classification as:
	\begin{definition}
		\label{def:comp_sym}
		A $\mathcal{G}$-action on $\mathcal{X}$ is a compatible symmetry for binary classification of classes sampled from two probability densities $P_0(\mathbf{X})$ and $P_1(\mathbf{X})_1$ with support $\mathcal{D}\subseteq\mathcal{X}$ if its acts faithfully on $\mathcal{D}$ and for any orbit $\Omega^\mathcal{X}(\mathbf{I})\subset\mathcal{X}$, the intersection $\mathcal{D}\cap\Omega^\mathcal{X}(\mathbf{I})\subset\mathcal{X}$ is either empty or it is contained in a single fibre of the likelihood ratio $\lambda(\mathbf{X})$.    
	\end{definition}	
	The requirement of a faithful group action naturally excludes the trivial action of any given group as a correct symmetry and we will exclude non-faithful group actions in our analysis. 	Additionally, the definition's negation naturally leads to an incompatible symmetry for binary classification, i.e. there is at least one $\mathbf{X}\in\mathcal{D}$ and $g\in\mathcal{G}$ such that $\lambda(\mathbf{A}_\mathcal{X}(g,\mathbf{X}))\neq\lambda(\mathbf{X})$ when $\mathbf{A}_\mathcal{X}(g,\mathbf{X})\in\mathcal{D}$. Thus, this naturally translates to the following self-evident observation. 
	\begin{observation}
		A $\mathcal{G}$-action on $\mathcal{X}$ is compatible with the likelihood ratio $\lambda(\mathbf{X})$ if and only if for every $\mathbf{X}\in\mathcal{D}$, $g\in\mathcal{G}$ such that $\mathbf{X}'=\mathbf{A}_\mathcal{X}(g,\mathbf{X})\in\mathcal{D}$ we have $\lambda(\mathbf{X}')=\lambda(\mathbf{X})$. 
	\end{observation}
Thus, any compatible group action need not have its entire orbit in $\mathcal{D}$, since the requirement of equality is mandated only  when the group action moves points within the support $\mathcal{D}$. This leaves room for equivariance under groups that contain the ``largest" symmetry of the likelihood ratio as a proper subgroup as long as the additional group elements take any $\mathbf{X}$ outside of $\mathcal{D}$. Before that, we can operationalise this observation to define the notion of a closed symmetry and a maximal symmetry of a binary classification task. 
\begin{definition}
	\label{def:closed_sym} 
	A compatible $\mathcal{G}_c$-action, say $\mathbf{A}_c(g,\mathbf{X})$ is a closed symmetry of a likelihood ratio if one can find at least one orbit of the action in $\mathcal{D}$, i.e $\Omega^\mathcal{X}(\mathbf{I}_0)\subset\mathcal{D}$ for some orbit $\Omega^\mathcal{X}(\mathbf{I}_0)$ of   $\mathbf{A}_c(g,\mathbf{X})$ in $\mathcal{X}$.  
\end{definition}  
	\begin{definition}
		\label{def:max_sym} 
		A $\mathcal{G}_0$-action, say $\mathbf{A}_0(g,\mathbf{X})$, is a maximal symmetry of some likelihood ratio $\lambda(\mathbf{x})$ if it is a compatible symmetry, and for any $\mathcal{G}$-action $\mathbf{\tilde{A}}_\mathcal{X}(g,\mathbf{X})$ that reduces to $\mathbf{A}_0$ when restricted to $\mathcal{G}_0$ as a proper subgroup, we have  $\mathbf{\tilde{A}}_\mathcal{X}(\tilde{g},\mathbf{X})\notin\mathcal{D}$ for every $\tilde{g}\in\mathcal{G}\setminus\mathcal{G}_0$ and $\mathbf{X}\in\mathcal{D}$. 
	\end{definition} 
 Note that since we require faithfulness for compatibility, it is not mandatory for every likelihood ratio to have a compatible symmetry resulting in the absence of closed or maximal symmetries. Nevertheless, if $\mathcal{G}_c$ or $\mathcal{G}_0$ exists, they characterise co-domain actions which maintain compatible fibre decomposition with the likelihood ratio.

\subsection{Characterising compatible co-domain group actions}  
 Utilising closed and maximal symmetries, we now characterise families of equivariant functions that will maintain a compatible fibre decomposition with the likelihood ratio.  First, consider that we have closed symmetry of the likelihood ratio. If any $\mathcal{G}$-action on the domain can be obtained by some restriction of $\mathbf{A}_c$ to some subgroup (not necessarily proper) of $\mathcal{G}_c$, then every group element $g\in\mathcal{G}$ has some  $\mathbf{X}\in\mathcal{D}$ such that $\lambda(\mathbf{X})=\lambda(\mathbf{A}_\mathcal{X}(g,\mathbf{X}))\neq0$. Thus, the corresponding action on the co-domain $\mathbf{A}_\mathcal{H}$ must act trivially for any element $g\in\mathcal{G}$ since otherwise $f(\mathbf{A}_\mathcal{X}(g,\mathbf{X}))\neq f(\mathbf{X})$ for any function $f(\mathbf{X})$ and $\mathbf{X}\in\mathcal{D}$. This can be captured as the following observation: 
	 \begin{observation}  
	 	\label{obs:no_go_co_domain} 
	 If $\mathbf{A}_\mathcal{X}$ is a compatible action of a group $\mathcal{G}\subseteq\mathcal{G}_c$ that is obtained by the  restriction of a closed action $\mathbf{A}_c$ to $\mathcal{G}$, then a function $f:\mathcal{D}\to\mathcal{H}$ that is equivariant under any action $\mathbf{A}_\mathcal{H}$ that acts non-trivially on the image $\im(f)\subseteq\mathcal{H}$ will have an incompatible fibre decomposition with the likelihood ratio $\lambda(\mathbf{X})$ in $\mathcal{D}$. 	
	 \end{observation}

	 Since the fibre decomposition is function specific, the restriction on the co-domain action is restricted to its image. However, from the perspective of function approximation by a neural network, the image changes dynamically during training which therefore forbids non-trivial action on the co-domain. Thus, under practical considerations, characterising a closed action $\mathbf{A}_c$ of some group $\mathcal{G}_c$ forbids any non-trivial equivariance on the co-domain that is obtained by some restriction to a subgroup of $\mathcal{G}_0$.

As we have assumed that compatible group actions need not have an entire orbit in the support $\mathcal{D}$, there can be situations where there are no closed symmetries even when there are compatible ones. However, there is some maximal symmetry when there are compatible ones.
Consider any compatible $\mathcal{G}$-action that reduces to the maximal action $\mathbf{A}_0$ under a restriction to $\mathcal{G}_0$ as a proper subgroup. If the two hypotheses are exhaustive (i.e. simple) and correct, no data can fall outside $\mathcal{D}$ even when the group action $\mathbf{A}_\mathcal{X}$ can take observed data outside under some transformation. Thus, if any element $\tilde{g}\in\mathcal{G}\setminus\mathcal{G}_0$ takes every point in $\mathcal{D}$ outside of it, then the $\mathcal{G}$-action is compatible under definition~\ref{def:comp_sym}. To have a compatible fibre decomposition in this case, it is sufficient for the action $\mathbf{A}_\mathcal{H}$ to act trivially for all $g\in\mathcal{G}_0$. Thus, we state this as an observation:

\begin{observation}       
	\label{obs:equi_opt} 
	If $\mathcal{G}_0$ is a proper subgroup of $\mathcal{G}$, and the restriction of $\mathbf{A}_\mathcal{X}$ to $\mathcal{G}_0$ reproduces $\mathbf{A}_0$ then equivariant functions $f:\mathcal{D}\to\mathcal{H}$ under the codomain action $\mathbf{A}_\mathcal{H}$ have compatible fibre decomposition with $\lambda(\mathbf{X})$ in $\mathcal{D}$ if for any $\mathbf{X}\in\mathcal{D}$ and $\tilde{g}\in\mathcal{G}\setminus\mathcal{G}_0$, we have $\mathbf{A}_\mathcal{X}(\tilde{g},\mathbf{X})\notin\mathcal{D}$, and $\mathbf{A}_\mathcal{H}(g_0,\mathbf{H})=\mathbf{H}$ for any $\mathbf{H}\in\im(f)$ for $g_0\in\mathcal{G}_0$.   
\end{observation}  

Evidently, there is no requirement on the action $\mathbf{A}_\mathcal{H}(\tilde{g},\mathbf{H})$ for $\tilde{g}\in\mathcal{G}\setminus\mathcal{G}_0$ since by construction no two elements in $\mathcal{D}$ are connected by such elements. Nevertheless, these can effect function approximation when one extrapolates to data that is not strictly contained in the support $\mathcal{D}$. This could happen, for instance, in the presence of noise, or when the assumed probabilities do not exactly describe the true sampling distributions. As we are neither analysing the effect of noise nor the behaviour of equivariant models on data with different characteristics, we will restrict our discussions to invariant models since under the two observations above, one requires a trivial action on the co-domain.

\subsection{Practical consequences}    
 In machine learning literature, one has various notions of correctness~\cite{wang2023the,NEURIPS2023_7dc7793c} and optimality~\cite{petrache2023approximationgeneralization}. Nevertheless, the observations made on this work provides a complementary understanding of the theory of correct, incorrect, and extrinsic equivariance developed in ref.~\cite{wang2023the,NEURIPS2023_7dc7793c} with regards to binary classification. Explicitly, while the notion of correct and incorrect symmetries are tied to the support $\mathcal{D}$ and the target function, extrinsic equivariance where the group action takes elements in $\mathcal{D}$ to regions where there is zero probability can either be beneficial or detrimental. From observation~\ref{obs:equi_opt}, we see that the utility of such extrinsic equivariance for binary classification tasks (or any general group invariant task) can be inferred by the relation of the stabilisers of the co-domain's action to the group $\mathcal{G}_0$.

In fundamental applications, our findings provide a rigorous basis for intelligent design of architectures based on the likelihood ratio. Especially for new-physics searches in particle colliders like the Large Hadron Collider (LHC), except for the process of hadronisation, one has a concrete and verified first-principles understanding of the sampling probability distributions from the underlying Lagrangian. These are utilised in the rich ecosystem of simulation tools generally based on Monte-Carlo methods. Often due to this  rich knowledge, the symmetries assumed in the forward process is utilised in the design of architectures. However, it has been found that these symmetries need to be explicitly reduced based on the inverse problem of statistical inference through likelihoods. A prototypical example is the classification of jets (collimated sprays of particles) to infer how they were produced in the initial collision. While the need for a strictly reduced continuous symmetry contained in the Lorentz group O(1,3) has been identified, state-of-the-art architectures generally consider permutation invariance over the entire jet constituents even when there is differentiating information like charge, and the type of quark that lead to the jet. 

For instance, to distinguish a top jet from one originating from a quark or a gluon, information of the bottom quark provides a reduced permutation symmetry amongst different constituents. This difference arises from the initial stages of the physical evolution described by perturbative dynamics of the Lagrangian. A top quark decays into a bottom quark and a $W$ boson, and a top jet must always contain a bottom quark. On the other hand, not all quarks and gluon jets consist of bottom quark but can arise if the initial particle was a bottom quark or if a gluon splits to an bottom-antibottom pair during the perturbative parton shower evolution. Therefore, one requires a reduced permutation symmetry to utilise this information efficiently.  Explicitly, the set of constituents of the jet $J=\{p_1,p_2,....,p_{n}\}$ partitions into two mutually exclusive subsets $S_1$ and $S_2$, where the constituents in (say) $S_1$ originate from the bottom quark, and those in $S_2$ from the $W$ boson. The invariant mass-squared $m^2= p^\mu.p_\nu$ of the $S_2$ set should fall near that of the $W$ boson mass for a true top jet. Additionally, if any combination of four vectors between the two sets falls near the $W$ boson's mass it is not likely to originate from a top quark. Thus, this information is available when we have permutation invariants that do not mix the two sets.

Another example of the suitability of a reduced permutation symmetry is in quark-gluon classification, where due to the difference in colour charge and electric charge, there is a difference in the particle content of different species of particles. However, there is no first-principles understanding of since the low energy fragmentation process has large non-perturbative effects. Nevertheless, in terms of raw performance, there is a theoretical impetus to utilise reduced permutation symmetries based on the particle content of the jets.

\section{Experiments}
\subsection{Toy scenario} 
We first consider a toy scenario in 3D point cloud classification to verify the optimality of different continuous symmetries. For all experiments, we add 3D normally distributed noise of diagonal covariance 0.3 to the cartesian coordinate representation.  We consider groups the groups E(3), O(3) and O(2) with input vector actions on $\mathbb{R}^3$, where the O(2) action acts along the $z$-axis with the modified metric signature $(1,1,0)$, in the evaluation of the norm and the inner product. From the probabilities specified below, we sample 30 points to construct a data sample for input to the neural network.  
 The feature extraction module is based on group equivariant message passing operations of the respective groups after which we take the mean readout of the sample of the invariant representation of the final message passing operation as inputs to a dense classifier network.
Additional details of the network architecture and training can be found in the appendix~\ref{app:arch}. 
We train the same architecture from random initialisation ten times and compute the relevant mean and standard deviation.  For both cases, the  data are randomly generated using the  \texttt{random.default\_rng} implemented in \textsc{NumPy} (v1.26.4 for the truncated normal case and v1.23.5 for the uniform case) with fixed seeds. For each class we independently sample 100k training, 40k validation, and 100k test samples.

\paragraph{Uniform:} We take a simple example of classifying a point cloud of a hollow cylinder and a sphere where the points are uniformly sampled from their embedding in 3D. The sphere and the cylinder have a unit radius, and we align the cylinder to the $z$-axis in the range $(-1,1)$. Due to the added noise, this is a simple yet non-trivial point cloud binary classification scenario where the largest symmetry of the probability distribution for the sphere is O(3), and that of the cylinder is O(2) with the axis fixed to the $z$-axis. Therefore, since O(2) is a subgroup of O(3), the largest group action under which the likelihood is invariant under the O(2) action along the $z$-axis.

\paragraph{Truncated Normal:} To understand the generalisation capabilities in scenarios where the symmetry lies in the probability itself and not their support, we consider a truncated ball and a cylinder. The radius for the ball and the cylinder follow a truncated normal distribution in the range $(4,6)$ centred at five and unit standard deviation. The azimuthal angle for both cases follows a uniform distribution in the range $(-\pi/4,\pi/4)$. For the sphere, we uniformly sample the polar angle in the range $(\pi/4,3\pi/4)$ while for the cylinder, we sample $z$-coordinates uniformly in the range $(5\cos 3\pi/4,5\cos\pi/4)$. Therefore, with the added noise of 0.3 standard deviations in cartesian coordinate representation, the probabilities have the same support on $\mathbb{R}^3$ with the same underlying (approximate) symmetries as the previous case.

\paragraph{Results:} For each of the best models per training, the background rejection: $\bar{\epsilon}_0(\epsilon_1)=1-\epsilon_0(\epsilon_1)$ at signal acceptance $\epsilon_1=0.95$, and the AUC  is evaluated on the test dataset, and we show its mean and standard deviation in table~\ref{tab:r95_both}. 
Comparing the different architectures, we find that the largest group E(3), leads to the poorest validation error for either data size, with a slight improvement for O(3) but still lower than the best-performing O(2). While the difference is significantly reduced for the equivariant case, the larger groups E(3) and O(3) have not been able to match the performance of O(2) equivariance, which confirms the compatibility of the smaller group O(2), and incompatibility of both O(3) and E(3) based on the underlying distributions.

\begin{table}[]
	\centering 
	\scalebox{.85}{
		\begin{tabular}{ccccc}
			\toprule 
			Sym. &\multicolumn{2}{c}{Uniform}&\multicolumn{2}{c}{Normal}\\

			&AUC&  $\bar{\epsilon}_0(\epsilon_1=0.95)$  & AUC & $\bar{\epsilon}_0(\epsilon_1=0.95)$\\
			\midrule  
			E(3) & 0.974$\pm$ 0.001&0.868$\pm$0.007& 0.920$\pm$ 0.003& 0.652$\pm$0.010\\ 
			O(3) &0.981$\pm$ 0.000&  0.905$\pm$0.002 &0.998$\pm$ 0.000&  0.994$\pm$0.000 \\ 
			O(2) & \textbf{1.000$\pm$ 0.000}&  \textbf{1.000$\pm$0.000}& \textbf{0.999$\pm$ 0.000}&   \textbf{0.999$\pm$0.000}\\ 
			
		\end{tabular}
	}
	\caption{ The mean and standard deviation of the background rejection at 95\% signal acceptance $\bar{\epsilon}_0(\epsilon_1=0.95)=1-\epsilon_0(\epsilon_1=0.95)$ and the AUC over ten training runs for the two toy scenarios with added noise.  The best performing values are highlighted in bold.}
	\label{tab:r95_both} 
\end{table}
\subsection{Jet classification}  

As a concrete application to fundamental applications, we consider jet classification at the Large Hadron Collider and the effect of different symmetry groups. The general purpose detectors at LHC collects huge amounts of highly granular data based on the different components of the detector. These contain information about the four-momentum $p_\mu\equiv(E,p_x,p_y,p_z)$, the charge, and the type of particles. Based on the likelihood ratio constructed out of the matrix-elements of a process, the largest continuous symmetry  corresponds to the longitudinal invariant subgroup O(1,1)$_l\otimes $O(2)$_t$ where the O(1,1)$_l$ group acts on $(E,p_z)$ and the $O(2)_l$ group acts on $(p_y,p_x)$. This is a proper subgroup of the Lorentz group O(1,3) acting on the four vectors $p_\mu$. Moreover, since there are group elements in O(1,3) (for instance transverse boosts) that do not belong to the longitudinal boost invariant subgroup that can change the likelihoods due to the finite detector resolutions, the Lorentz group is an incompatible symmetry.  However, the longitudinal subgroup is not the maximal group $\mathcal{G}_0$ since one has to consider permutation symmetries within the constituents. This depends on the particular tagging scenario and selection criteria. For both scenarios, we consider the combination of the two continuous groups with two permutation symmetries: one over the entire final state particles $S_n$, and the other cover a product subgroup $\otimes_{t=1}^{N} S_{m_t}$ with .  This is implemented by independent mean readout operations over each subsets and concatenating them to form a representation that is invariant under the reduced group $\sum_{t=1}^N\;m_t=n$. Since invariants of the larger $S_n$ groups are themselves invariant under this reduced action to a subgroup, we also add the readout over the entire constituents for those architectures with the product permutation symmetry. Additional details of the architectures and training are given in  the appendix~\ref{app:arch}.

\paragraph{Top tagging:}  As already explained before, the bottom quark's information is crucial in correctly distinguishing a top jet from a quark/gluon jet, especially when one considers the combinatorial masses of the constituents. However, publicly available top-tagging datasets neither have the bottom quark information nor do they put requirements of the jet mass falling near the mass of the top quark $m_t=173$ GeV. Therefore, we consider the dataset utilised in ref.~\cite{Ngairangbam:2023cps} that contains the bottom quark information. This dataset consists of simulated jets divided into two labels where one class originates from a en energetic quark or a gluon while the other originates from the decay of a top quark. Each jet falls has transverse momentum $p^{\text{Jet}}_T\in(400,500)$ GeV. The constituents of the jets are reclustered into microjets (i.e. smaller jets) with $p_T^{m}> 5$ GeV, with a jet-dependent radius $R(I_J)=0.1+I_J\;0.01$ where for each jet, $I_J\in\{0,1,2,...\}$ is iteratively increased from zero until the clustering produces at most eight microjets. These microjets additionally have a B-tagging information by utilising the simulation-level identity of their constituents. For additional details, we refer interested readers to section~5.1 of ref.~\cite{Ngairangbam:2023cps}. Since, the bottom quark information directly relates to the invariant mass, we put an additional selection criterion on the jet mass to fall near $m_t$, i.e. we have some positive quantity $\Delta \Gamma$ such that the invariant mass of the jet $m_J$ falls in the interval $[m_t-\Delta \Gamma,m_t+\Delta\Gamma]$. To diagnose the dependence of the discrete symmetry on this mass window, we consider two mass windows: $\Delta \Gamma=30$ GeV and $\Delta\Gamma=50$. After this selection criterion, we use 150k and 200k samples-per-class, respectively, for the two values. Out of the total, half of the samples are used for training the models, with the remaining half equally divided for validation and testing.  

\paragraph{Quark/Gluon tagging:} We use the publicly available dataset~\cite{komiske_2019_3164691} to analyse the effects of the different symmetries in quark-gluon tagging. This simulated dataset consists of labelled quark and gluon jets with transverse momentum in the range $[500,550]$ GeV. Each sample consist of the four-vector of the jet's massless constituents and the particle identity~\cite{ParticleDataGroup:2024cfk}. We use the particle identity information to divide the constituents into four types: (i) negatively charged mesons and leptons, (ii) positively charged mesons and leptons, (iii) neutral mesons and the photon, and (iv) all other hadrons. We utilise 200k samples from each class with the entire 400k divided into 200k training, 100k validation, and 100k test samples.

\begin{table}
	\centering 
	\begin{tabular}{ccc}
		\toprule
		Task&Particle type & Value \\
		\midrule 
		
		\multirow{4}{*}{Quark/Gluon tagging}&Negatively charged mesons and leptons & 1\\ 
		&Positively charged mesons and leptons & 2\\ &Neutral mesons and the photon& 3\\ 		 
		&Rest & 0\\
		\midrule 
		\multirow{2}{*}{Top tagging}&Bottom & 1\\
		&Rest & 0\\
		\bottomrule  
	\end{tabular} 
	\caption{Scalar node attribute for particle type utilised in the jet classification experiments.
	}
	\label{tab:part_embed}
\end{table}

\paragraph{Input representation:} For both scenarios, we construct complete graphs out of the constituent particles with scalar and vector node representations. For models consisting of the O(1,3) symmetry, we use only the four vector $p_\mu$ of each particle, while the scalar attribute consists of an integer that signifies the particle type. For both scenarios, this is given in table~\ref{tab:part_embed}. For the $S_n$ invariant models this feature serves as a means to provide information of the different particle types where the graph readout can only retain invariant information over the entire jet's particle content. In contrast, smaller permutation symmetries can retain information about the particle types separately. For the O(1,1)$_l\otimes$O(2)$_t$ models, we construct the node feature $\mathbf{h}_i=(\phi_i,\log p^t_i,T_i)$, where $T_i$ denotes the particle type given in table~\ref{tab:part_embed}, and the scalar edge features $\mathbf{e}_{ij}=(\log (p^t_ip^t_j),\Delta y_{ij},\Delta\phi_{ij},\Delta R_{ij})$.

\paragraph{Results:} The averaged AUC over five training instances and its standard deviation  are given in Tables~\ref{tab:top_exp} and ~\ref{tab:qg_exp} for the top-tagging, and quark-gluon classification scenarios, respectively. Amongst all cases, we see that the combination of the longitudinal invariant and reduced permutation invariance perform the best. The worst performing architecture corresponds to one where both the continuous and discrete symmetries are both incorrect. It is important to note that the input node representations contain a feature which distinguishes the different particle types for all architectures, which however, does not lead to breaking of the underlying larger permutation symmetry in $S_n$ models as they can only learn invariants built out of entire particle content. Moreover, we see that the effect of the bottom quark information for top tagging (when one reduces the permutation symmetry), has a larger effect when $\Delta \Gamma=30$ GeV, i.e. when the background jets have similar mass range to the top jet. Nevertheless, for $\Delta\Gamma=50$ GeV, the correct discrete symmetry enhances the performance even when the continuous symmetry is incorrect.  Hence, we see that  knowledge of the fibre decomposition coupled with domain knowledge can help guide the design 
of group symmetries that are appropriate for the task.

\begin{table} 
	\centering 
	\scalebox{0.85}{
	\begin{tabular}{cccccc}
		\toprule
		Cont. Sym. & Discrete Sym. & \multicolumn{2}{c}{$\Delta \Gamma=30$ GeV} & \multicolumn{2}{c}{$\Delta \Gamma=50$ GeV}\\
		&& AUC& $\bar{\epsilon}_0(\epsilon_1=0.95)$ & AUC& $\bar{\epsilon}_0(\epsilon_1=0.95)$ \\
		\midrule
				\multirow{2}{*}{O(1,3)} & $S_n$ & 0.972$\pm$0.001 &0.891$\pm$0.008 &0.979$\pm$0.000 &0.915$\pm$0.002\\
		& $\otimes_{t}\; S_{m_t}$ & 0.977$\pm$0.000&0.933$\pm$0.002 & 0.983$\pm$0.000 &0.937$\pm$0.001\\		
		\midrule 
		\multirow{2}{*}{O(1,1)$_{l}\otimes$ O(2)$_{t}$} & $S_n$ & 0.977$\pm$0.001&0.933$\pm$0.006 &0.984$\pm$0.001&0.941$\pm$0.003  \\
		& $\otimes_{t}\; S_{m_t}$ & \textbf{0.978$\pm$0.000} & \textbf{0.938$\pm$0.002} &\textbf{0.984$\pm$0.000} &\textbf{0.942$\pm$0.002}\\
		\bottomrule
	\end{tabular}
}
	\caption{The mean and standard deviation of the background rejection at 95\% signal acceptance $\bar{\epsilon}_0(\epsilon_1=0.95)=1-\epsilon_0(\epsilon_1=0.95)$ and the AUC over five training runs for top tagging datasets with different mass windows $\Delta\Gamma$.  The best performing values are highlighted in bold.}
	\label{tab:top_exp}
\end{table}

\begin{table} 
	\centering 
	\scalebox{0.85}{
\begin{tabular}{cccc}
	\toprule
	Cont. Sym. & Discrete Sym. & AUC & $\bar{\epsilon}_0(\epsilon_1=0.95)$\\
	\midrule 
\multirow{2}{*}{O(1,3)} & $S_n$ & 0.875$\pm$0.000&0.450$\pm$0.002 \\
& $\otimes_{t}\; S_{m_t}$ & 0.879$\pm$0.001& 0.463$\pm$0.002\\
	\midrule
	
	\multirow{2}{*}{O(1,1)$_{l}\otimes$ O(2)$_{t}$} & $S_n$ & 0.900$\pm$0.003 &0.531$\pm$0.010\\
	& $\otimes_{t}\; S_{m_t}$ & \textbf{0.901$\pm$0.001}& \textbf{0.532$\pm$0.005}\\
	\bottomrule
\end{tabular}
}
\caption{The mean and standard deviation of the background rejection at 95\% signal acceptance $\bar{\epsilon}_0(\epsilon_1=0.95)=1-\epsilon_0(\epsilon_1=0.95)$ and the AUC over five training runs for quark vs gluon classification. The best performing values are highlighted in bold.}
\label{tab:qg_exp}
\end{table}

\section{Conclusions} 

 In this work, we have presented a novel framework for characterising optimal group symmetries in binary classification tasks. Our theoretical analysis and experimental results demonstrate that the optimal selection of group symmetries, aligned with the intrinsic properties of the underlying data distribution, is critical for enhancing both generalisation and sample efficiency.

Leveraging the Neyman-Pearson optimality of monotonic invertible functions of the likelihood ratio, we argue that any binary classification task has an associated fibre decomposition on the support of the two probabilities. Thus compatibility of different symmetries with the likelihood ratio can be ascertained by the \emph{minimal fibres} that any group invariance imposes on the domain via its orbits.  In this sense, we identify the role of the maximal symmetry defined as the smallest group whose action on the domain has the largest minimal fibres decomposition compatible with the likelihood ratio on the support of the probabilities. Consequently, we find that when any group action on the domain results from a restriction of this maximal symmetry to some subgroup, any non-trivial group action on the co-domain (observation~\ref{obs:no_go_co_domain}) has incompatible fibre decompositions. On the other hand, we  characterise properties of group actions that reduces to this maximal symmetry under a proper subgroup restriction   which still maintains compatibility (observation~\ref{obs:equi_opt}). Thus our work provides a concrete basis for the design of optimal and symmetric architectures based on the underlying likelihood ratio.

We demonstrated a basic example of how the insights can be put into action considering a toy binary classification scenario. Furthermore, we demonstrated how the observations can help guide symmetric architecture design for jet tagging at the Large Hadron Collider. Concretely, we identified a still overlooked feature in current literature where one considers permutation invariance over all jet constituents as suboptimal. For the decay of heavy resonant particles like the top quark, numerical evidence suggests that such reduced permutation symmetries becomes more important when the jet mass goes closer to its mass peak.

Future work will focus on extending this framework to multi-class classification problems and exploring the utility of our approach in other domains where symmetries play a crucial role, such as in physics-based simulations and high-dimensional data analysis. Additionally, we aim to investigate further the impact of noise and other real-world factors on the performance of group equivariant architectures to enhance their robustness and applicability in practical scenarios.
\section*{Acknowledgements}
This work was funded by the STFC under grant ST/X003167/1.
\appendix

\section{Details of architecture and training} 
\label{app:arch} 
 
 \paragraph{Group equivariant message passing:} Our architecture is based on the group equivariant message passing operations of ref.~\cite{villar2021scalars} for the classical groups.
At the $l$-th stage of the message passing operation, the equivariant operations takes  the scalar node representations $\mathbf{h}^{(l)}_i$ and the vector node representation $\mathbf{x}^{(l)}_i$ via the following equations: 
\begin{subequations} 
	\label{eq:equiv_mpnn} 
	\begin{align}
		\label{eq:scalar_msg}
		\mathbf{m}^{(l+1)}_{ij}&=\Phi^{(l+1)}_e(\mathbf{h}^{(l)}_i,\mathbf{h}^{(l)}_j,\mathbf{e}_{ij}^{(l)})\quad,\\
		\label{eq:vec}
		\mathbf{x}_i^{(l+1)}&=\mathbf{x}^{(l)}_i+\sum_{j\in\mathcal{N}(i)} \; \mathbf{v}^{(l)}_{ij}\; \Phi^{(l+1)}_x(\mathbf{m}^{(l+1)}_{ij})\quad,\\
		\mathbf{m}^{(l+1)}_i&=\sum_{j\in\mathcal{N}(i)} \;\mathbf{m}^{(l+1)}_{ij}\quad,\\
		\mathbf{h}^{(l+1)}_i&=\Phi^{(l+1)}_h(\mathbf{h}^{(l)}_i,\mathbf{m}^{(l+1)}_i)  \quad. 
	\end{align}
\end{subequations}
Where $\mathbf{v}^{(l)}_{ij}=\mathbf{x}^{(l)}_i-\mathbf{x}^{(l)}_j$ for the inhomogeneous group E(3), and $\mathbf{v}^{(l)}_{i,j}= \mathbf{x}^{(l)}_j$ for the rest. For the toy experiments, the scalar edge attributes are $\mathbf{e}^{(l)}_{ij}=(|\mathbf{x}^{(l)}_i-\mathbf{x}^{(l)}_j|^2,\langle\mathbf{x}_i,\mathbf{x}_j\rangle)$ for the O(3) and O(2) groups while for the E(3) group  it is $\mathbf{e}^{(l)}_{ij}=|\mathbf{x}^{(l)}_i-\mathbf{x}^{(l)}_j|^2$.  For the jet classification tasks on four vectors $\mathbf{x}_i\equiv p_\mu=(E,p_x,p_y,p_z)$, we consider the Minkowski metric $\eta_{\mu\nu}=\text{diag}(-1,1,1,1)$ for the O(1,3) group, while for the O(1,1)$_l\otimes$O(2)$_t$ case, we additionally take  $\eta^l_{\mu\nu}=\text{diag}(-1,0,0,-1)$ and  $\eta^t_{\mu\nu}=\text{diag}(0,1,1,0)$. For each metric signature, we construct three two-body invariants: $s_{ij}=\langle\mathbf{x}_i+\mathbf{x}_j,\mathbf{x}_i+\mathbf{x}_j\rangle$, $d_{ij}=\langle\mathbf{x}_i-\mathbf{x}_j,\mathbf{x}_i-\mathbf{x}_j\rangle$, and $c_{ij}= \langle\mathbf{x}_i,\mathbf{x}_j\rangle$. To alleviate the exponentially separated scales present in parton showers, we use the function $R(e) = \text{sign}(e)\log(|e| + 1)$ for these three inner products. 
Thus, we have three and nine edge scalars for the O(1,3) and O(1,1)$_l\otimes$O(2)$_t$ groups, respectively. The functions $\Phi_e^{(l+1)}$, $\Phi^{(l+1)}_x$ and $\Phi^{(l+1)}_h$ are MultiLayer Perceptrons (MLPs), with $\Phi^{(l+1)}_x$'s output a single value made to fall in the open unit interval via the sigmoid activation function.

\paragraph{Hyperparameters:} For all architectures, we have a total of three equivariant message passing operations which uses ReLU activation in all input and hidden layers.  All MLPs have two hidden layers of sixty nodes for the toy experiments and one-hundred and twenty eight nodes for jet-classification.
The output of all $\Phi^{(l+1)}_e$ also correspond to the same dimensions as the hidden layers which unambiguously fixes the input dimensions of the other two functions.  While $\Phi^{(l+1)}_x$ has a sigmoid output activation, we do not apply any non-linear activation to the other outputs. We forego the vector update at the final layer and use only the  mean global representation of each scalar node representation as input to a classifier. For the reduced permutation symmetries, we concatenate the mean scalar representation over each factor in $\otimes_{t}\;S_{m_t}$.  The classifier network has a single output node with sigmoid activation.  With this set of hyperparameter values, the invariant models have 120k parameters while the equivariant ones have 130k parameters for the toy scenario. For jet classification with $S_n$ symmetry, the    O(1,1)$_l\otimes$O(2)$_t$ and O(1,3) models have  451k and 447k trainable parameters, respectively.  For the reduced permutation symmetries, the O(1,3) models have 480k and 512k parameters for the top tagging and quark gluon tagging scenarios, respectively, while for the O(1,1)$_l\otimes$O(2)$_t$ group they have and 484k and 516k parameters.   

\paragraph{Training:} All network training uses binary cross entropy loss and is trained ten times from random initialisation for hundred epochs. We use the \textsc{Adam} optimiser~\cite{DBLP:journals/corr/KingmaB14} with initial learning rate of 0.001, which decays by a factor of 0.5 if the validation loss has not improved over three epochs. We use \textsc{PyTorch } for all numerical experiments. For the truncated normal experiments, we use version 2.2.1 with  \textsc{cudatoolkit 11.8.0} while for the uniform scenario and jet classification we use version 1.12.1 with \textsc{cudatoolkit 11.3.1}. The models for the jet classification were implemented in \textsc{PyTorch-Geometric} version 2.2.0. All experiments were carried out on \textsc{NVIDIA A100} or \textsc{V100} GPUs. Training is either carried out on two homogeneous GPUs or on a single GPU with a total effective batch size of 300 samples for the toy scenario and 128 samples for jet classification.

\bibliographystyle{JHEP}
\bibliography{ref}

\end{document}